\documentclass[10pt,twocolumn,letterpaper]{article}

\usepackage{cvpr}
\usepackage{subfigure}
\usepackage{times}
\usepackage{epsfig}
\usepackage{graphicx}
\usepackage{amsmath,amsfonts,amsthm,amssymb}
\usepackage{multirow}

\makeatletter

\newcommand{\tabincell}[2]{\begin{tabular}{@{}#1@{}}#2\end{tabular}}

\newcommand{\Rmnum}[1]{\expandafter\@slowromancap\romannumeral #1@}

\newcommand{\conv}{\text{{\tt conv}}}

\newcommand{\trans}{\text{{\tt trans}}}
\newcommand{\model}{SiCNN}

\newcommand{\fig}[1]{Fig.~\ref{fig:#1}}
\newcommand{\tab}[1]{Table~\ref{tab:#1}}
\newcommand{\secc}[1]{Section~\ref{sec:#1}}

\def\f{{\bf f}}
\def\I{{\bf I}}
\def\S{{\bf S}}
\def\Q{{\bf Q}}

\usepackage{multirow}
\usepackage[pagebackref=true,breaklinks=true,letterpaper=true,colorlinks,bookmarks=false]{hyperref}

\cvprfinalcopy % *** Uncomment this line for the final submission

\def\cvprPaperID{1738} % *** Enter the CVPR Paper ID here

\ifcvprfinal\fi
\begin{document}

%%%%%%%%% TITLE
\title{Scale-Invariant Convolutional Neural Network}

\author{
Yichong Xu \\
Tsinghua University\\
\tt{\small xycking@163.com}\\
\and
Tianjun Xiao\\
Peking University\\
\tt{\small xiaotianjun@pku.edu.cn}\\
\and
Jiaxing Zhang\\
Microsoft Research Asia\\
\tt{\small jiaxz@microsoft.com}
\and
Kuiyuan Yang\\
Microsoft Research Asia\\
\tt{\small kuyang@microsoft.com}
\and
Zheng Zhang\\
NYU Shanghai\\
\tt{\small zz@nyu.edu}
}

\maketitle
%\thispagestyle{empty}

%%%%%%%%% ABSTRACT
\begin{abstract}
%While achieving near-human performance in various computer vision tasks, convolutional neural networks (CNN) are essentially not efficient for scale-invariance, which prevents it from  generalization to real-world object recognition. In this paper, we propose a scale-invariant convolutional neural network (\model), a model designed to incorporate multi-scale feature exaction and classification into the network structure. {\model} uses a multi-column architecture, with each column focusing on a particular scale. Unlike previous multi-column strategies, these columns share the same set of filter parameters by a scale transformation between them. Experimental results show that {\model} detects features in various scales, and the classification result exhibits more invariance against the scale change of objects. On CIFAR-10 dataset, Our mutli-column solution improves the classification accuracy of many deep models from traditional CNN to maxout CNN. {\model} is even more attracting on datasets with bigger scale variation.

Even though convolutional neural networks (CNN) has achieved near-human performance in various computer vision tasks, its ability to tolerate scale variations is limited. The popular practise is making the model bigger first, and then train it with data augmentation using extensive scale-jittering. In this paper, we propose a scale-invariant convolutional neural network (\model), a model designed to incorporate multi-scale feature exaction and classification into the network structure. {\model} uses a multi-column architecture, with each column focusing on a particular scale. Unlike previous multi-column strategies, these columns share the same set of filter parameters by a scale transformation among them. This design deals with scale variation without blowing up the model size. Experimental results show that {\model} detects features at various scales, and the classification result exhibits strong robustness against object scale variations.
%On CIFAR-10 dataset, {\model}  improves the classification accuracy of many deep models from traditional CNN to maxout CNN. {\model} is even more attracting on datasets with bigger scale variation.
\end{abstract}

%%%%%%%%% BODY TEXT
\section{Introduction}

Many classical computer vision tasks have enjoyed a great breakthrough, primarily due to the large amount of training data and the application of deep convolution neural networks (CNN)~\cite{krizhevsky2012imagenet}. In the most recent ILSVRC 2014 competition~\cite{russakovsky2014imagenet}, CNN-based solutions have achieved near-human accuracies in image classification, localization and detection tasks ~\cite{simonyan2014very,szegedy2014going}.

Accompanying this progress are studies trying to understand what CNN has learnt internally and what contribute to its success~\cite{erhan2009visualizing,simonyan2013deep,zeiler2013visualizing}. By design, layers within the network have progressively larger receptive field sizes, allowing them to learn more complex features. Another key point is the shift-invariance property, that a pattern in the input can be recognized regardless of its position~\cite{lecun1995convolutional}. Pooling layers contribute resilience to slight deformation as well small scale change~\cite{scherer2010evaluation}.

However, it is evident that CNN deals with shift-variance far better than scale-invariance~\cite{gong2014multi}. Not dealing with scale-invariance well poses a direct conflict to the design philosophy of CNN, in that higher layers may see and thus captures features of certain plain patterns simply because they are larger at the input, not because they are more complex. In other words, there is no alignments between in the position of a filter and the complexity it captures. What is more, there are other invariance that CNN does not deal with internally. Examples include rotations and flips (since features of natural objects are mostly symmetric).

A brutal force solution would be to make the network larger by introducing more filters to cope with scale variations of the same feature, accompanied by scale-jittering the input images, often by order of magnitude. This is, in fact, the popular practice today \cite{ciresan2012multi,krizhevsky2012imagenet,simonyan2014very}. It is true even for proposals that directly deal with this problem.
For example, ~\cite{gong2014multi} drives the CNN with crops of different size and positions with three differnt scales, and then uses VLAD pooling to produce a feature summary of the patches.

We explore a radically different approach that is also simple. Observing that filters that detect the same pattern but with different scales bear strong relationship, we adopt a multi-column design and designate each column to specialize on certain scales. We call our system \model \; (Scale-invariant CNN). Unlike a conventional multi-column CNN, filters in \model \; are strongly regulated among columns. The goal is to make the network resilient to scale variance without blowing up number of free parameters, and thus reduce the need of jittering the input.
% ~\cite{ciresan2012multi}, CNN Unlike the methods above, our method involves a multi-column framework which is similar to ~\cite{ciresan2012multi}. We use several columns of a similar struture in our network, and then summarize the results of different columns into the softmax output. Unlike in ~\cite{ciresan2012multi}, we use the same training data for all columns of the network, and the weights in all columns are tied on a same set of parameters: The convolution layers has different size of receptive field across the column and each column is designed to focus on a particular scale. Compared with ~\cite{gong2014multi}, our method exclusively uses the whole original image and all the invariance is gained inside the network, which reduces the training and testing cost dramatically.

We performed detailed analysis and verified that {\model}  exhibits the desired behavior. For example, the column that deals with larger scale is indeed activated by input patterns with the larger scaling factor, and the system as a whole becomes less sensitive to scale variance. On unaugmented CIFAR-10 dataset ~\cite{krizhevsky2009learning}, our method produces the best result among all previous works using a single CNN and a simple softmax classifier, and is complementary to other techniques that improve the performance. Our model increases training cost linear to number of columns, but we find that incremental refinement can dramatically reduce the cost without compromising the performance significantly.

The rest of the paper is organized as follows. \secc{method} presents \model, covering the high-level intuition, the mathematical foundation and the architecture. Detailed analysis and results are presented in \secc{exprRes}, and we conclude in \secc{conclude}.
%\TODO{might do a discussion}
%We cover related work in Section 2 and then describe the pipeline utilizing visual attentions for fine-grained classification in Section 3. Detailed performance study and analysis are covered in Section 4. We discuss what we learned, future work and conclusion in Section 5.

%For similar works, the idea of using different size of receptive field in a same layer is applied in ~\cite{szegedy2014going} to make the net sparse enough. We didn't find results which uses a similar way of tying weights as in our model.

%\input{related.tex}
\section{Model} \label{sec:method}

Consider the case of classifying objects that have only one \emph{canonical} scale and the only free parameter is their positions. A stack of convolution filters can progressively build more complex hidden representations. These hidden representations are all \emph{invariant by shift}, meaning that the activations preserve the same pattern except that they are shifted. In other words, $\conv(\text{Shift}(I), f) \equiv \text{Shift}(\conv(I, f))$, for arbitrary image $I$ and filter $f$, and this relationship is upheld layer to layer. This makes the job of the classifier easy.

In the existing CNN architecture, dealing with multiple scales is jointly achieved by the pooling layers and the convolution layers. The convolution layer not only needs to learn different features but also their scaled variants into multiple feature maps. Units in the pairing pooling layer generate scale-invariance within their receptive fields, which help to save feature maps. This multi-scale solution leads to bigger model and, since filters are independently learned, the need of more training data. The popular practice is scale-jittering ~\cite{krizhevsky2012imagenet}.

Our idea is simple, and is inspired by the \emph{invariance-by-shift} property of the existing convolution layer. Just like CNN convolve a filter on different positions, we also "convolve" the filter on different scales. This is done by adding independent columns, each is a conventional CNN but ``specialized'' at detecting one scale. Crucially, the columns are strongly regularized such that the number of free parameters in the convolution layers stay the same. Thus, we inject scale-invariance into the model, requiring neither additional data augmentation nor increasing the model size.

In the followings, we first introduce our architecture, present the intuition and then give the concrete mathematical definition.

\subsection{Scale-Invariance Architecture}

\begin{figure}[htbp]
\center
\includegraphics[width=0.45\textwidth]{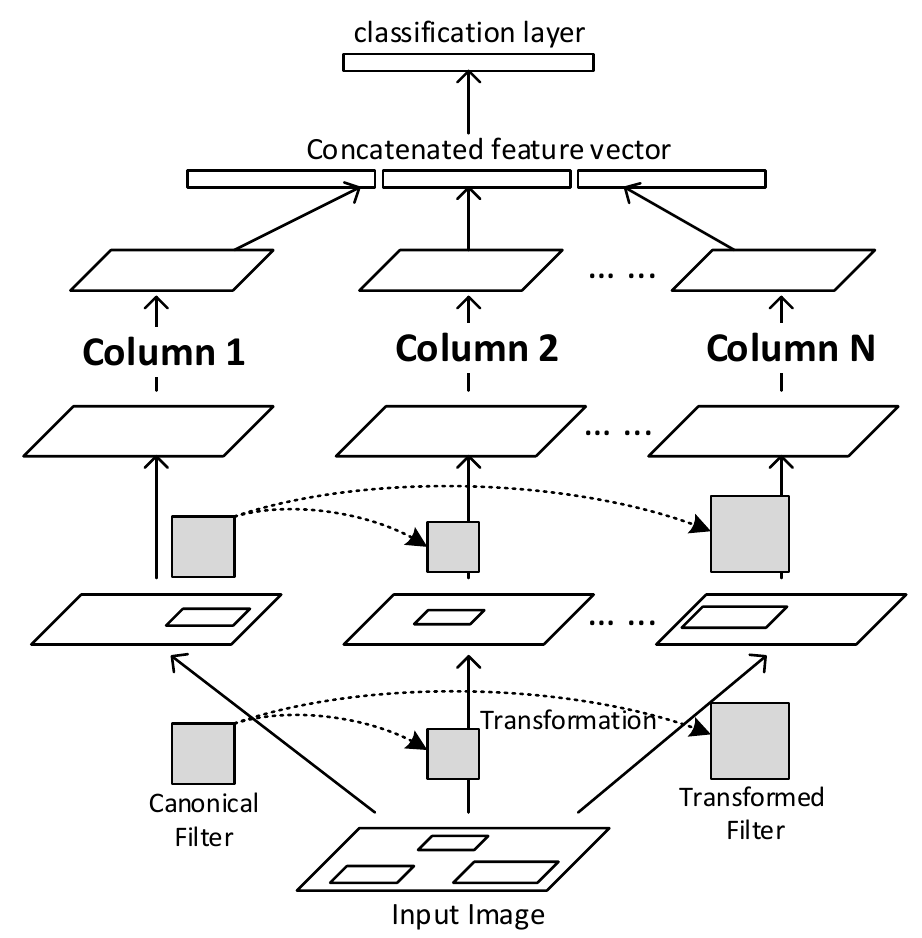}
\caption{Architecture of \model.}
\label{fig:architecture}
\end{figure}

\model \;uses multiple columns of convolutional stack with varying filter size to capture objects with unknown scales in input images. The architecture is illustrated in Figure~\ref{fig:architecture}. From bottom up, the input image is fed into all the columns. Each column has several convolutional layers with max-pooling. The key difference from conventional multi-column CNN is that, although these columns use different filter size, they still share a set of common parameters among their filters. A canonical column (Column 1 in Figure~\ref{fig:architecture}) keeps canonical filters in each layer. Other columns, which we call \emph{scale} columns, transform these canonical filters into their own filter. Collectively, a canonical filter and its transformed filters detect its pattern at different scales in multiple columns simultaneously.
Therefore, a single pattern with different scales trigger one or more columns.

In our architecture, we simply concatenate the top-layer feature maps from all the columns into a feature vector. The final classification layers (a softmax layer in the simplest case) take this feature vector as input.

\subsection{Filters in Multiple Scales}

\begin{figure}[htbp]
\center
\includegraphics[width=0.3\textwidth]{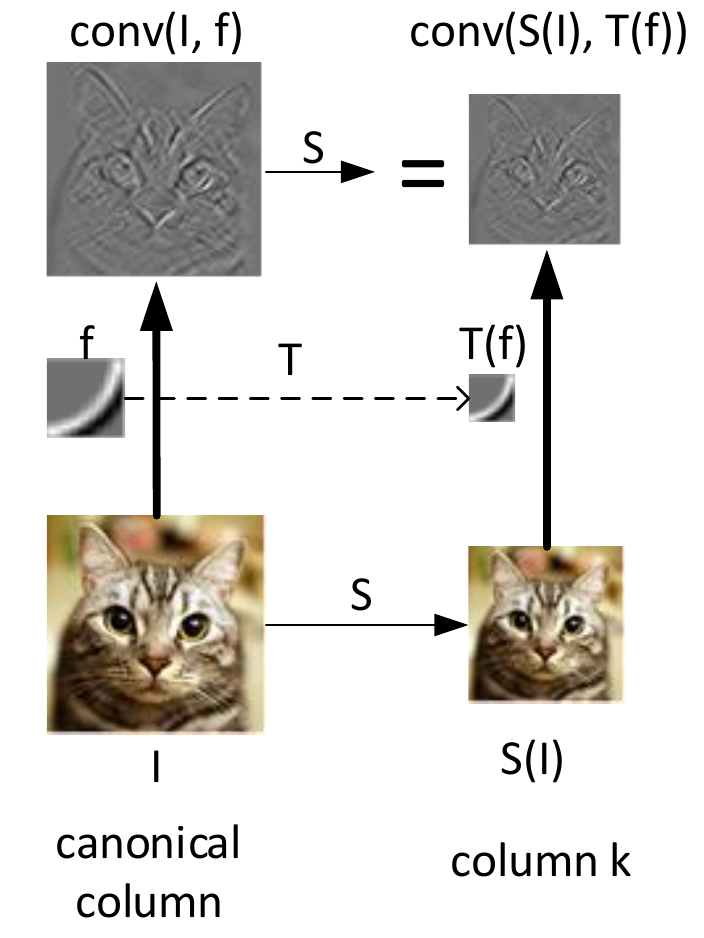}
\caption{The transformation from a canonical filter to another column. Best viewed in electronic form.}
\label{fig:filterTransformation}
\end{figure}

Filters that are transformed into different columns from the canonical filter capture the same pattern at different scales. We will discuss this transformation from canonical filters to other columns.

Consider a canonical filter $f$, which detects a pattern in image $I$ by convolution $\conv(I,f)$ (Figure~\ref{fig:filterTransformation}). When the image is scaled by a scaling operation $S$ to $S(I)$, we expect another column $k$ with transformed filter $T(f)$ to capture the same pattern instead. Thus, the column $k$ generates another convolution result $\conv(S(I), T(f))$. Just as invariance-by-shift, we require this convolution be equivalent with scaling from the convolution result in the canonical column. That is
\begin{align}
\conv(S(I),T(f)) = S(\conv(I,f)). \label{eq:invariance}
\end{align}
We call this property of filter as \emph{invariance-by-scaling}. Given a scaling $S$, we want to find the $T$ that satisfies Equation~\ref{eq:invariance} for any image $I$ and filter $f$.

\begin{figure}[htbp]
\center
\includegraphics[width=0.4\textwidth]{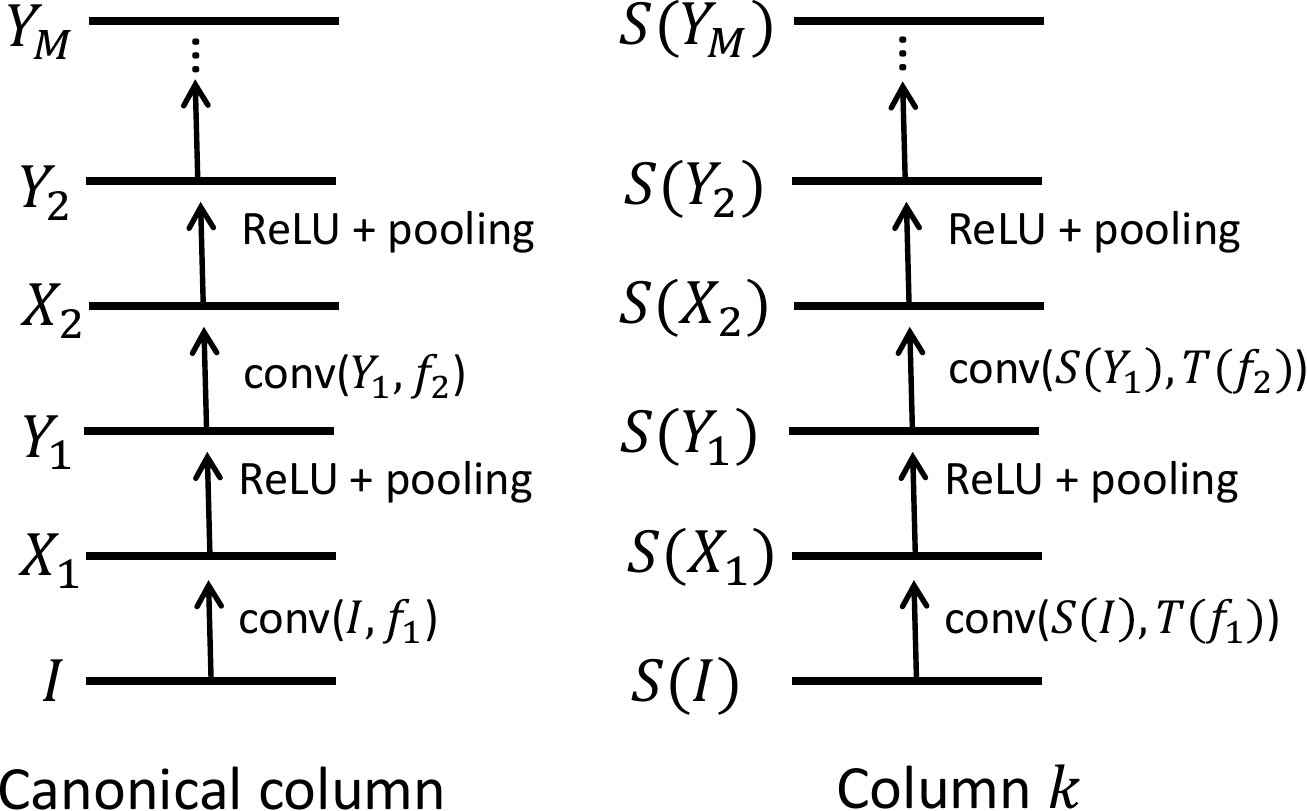}
\vspace{1 em}
\caption{Columns with multiple layers to capture patterns in different scales.}
\label{fig:multiLayer}
\end{figure}

The above discussion is for the first layer. However, it is easy to see that if the filter transformation $T$ in each layer satisfies Equation~\ref{eq:invariance}, then invariance-by-scaling property is preserved layer by layer till reaching the classification layer. In Figure~\ref{fig:multiLayer}, when input images $I$ and $S(I)$ with different scale are fed into the canonical column and the column $k$ separately, they generate $X_1$ and $S(X_1)$ respectively. The following ReLU nonlinear activation function and max-pooling keep this scale relationship and result in $Y_1$ and $S(Y_1)$. By recursively applying Equation~\ref{eq:invariance}, we know the top layers of these two columns also keep the same scale relationship: if canonical column generates $Y_M$ on input image $I$, the column $k$ generates $S(Y_M)$ on image $S(I)$.

If the object scale fits exactly one of the columns, then there is a perfect matching with the column outputting the highest responses. Otherwise, if the object scale just falls in between the scales of two neighboring columns, both the two columns will have relatively high responses. The concatenated feature vector at the end makes it possible for the classifier to do a linear combinations of responses from multiple columns to eliminate above variance.

\subsection{Filter Transformation}\label{sec:scalefilter}

By a vector representation of the image (concatenating all the rows or columns in matrix), scaling and convolution are all linear transformations. Given a canonical filter $f_c$ represented by a vector, we can solve the following equation derived from Equation~\ref{eq:invariance} to get the transformed filter $f_t$,
\begin{align}
\forall I, \conv(S(I),f_t) = S(\conv(I,f_c)). \label{eq:fc2fs}
\end{align}
Equation~\ref{eq:fc2fs} is a system of linear equations for $f_t$. However, such a system doesn't always have a valid solution because it has too many constraints (linear equations). To address this problem, we reduce $I$ to be of the same size as $f_c$, which makes the $\conv(I,f_c)$ produce only a single number.
%Such a method would also work for large images, since in large images $k_1$ also multiplies with small patches on the image.
Then, Equation~\ref{eq:fc2fs} turns into
\begin{align}
\forall \I, (\S \cdot \I)^T \cdot \f_t = \I^T \cdot \f_c, \label{eq:fc2fs_matrix}
\end{align}
where $\S$ is the scaling matrix, $\I$ is the vector representation of the image, and similar are $\f_c$ and $\f_t$.
It's easy to prove that Equation~\ref{eq:fc2fs_matrix} is equivalent to
\begin{align}
\S^T \cdot \f_t = \f_c, \label{eq:fc2fs_final}
\end{align}
We can solve Equation~\ref{eq:fc2fs_final} to obtain $\f_t$. However, in practice we can't always obtain an exact or unique $\f_t$ because $\S$ is not a square and invertible matrix. When $\S$ is a scaling-up matrix (\#rows $>$ \#columns), the equation have infinite number of solutions; when $\S$ is a scaling-down matrix (\#rows $<$ \#columns), the equation have no exact solutions.

%Subsampling methods can be seen as a linear combination of the original pixels. Suppose $Q$ is the transformation matrix for image of size same as $k_1$ to resize to same as $k_2$, i.e.,
%\[(S(I))^r=QI^r.\]
%Here $I^r$ means stretching the matrix $I$ to a column vector, in a column-first manner. Then we have the following theorem:\\
%
%\textbf{Theorem 4.} If $S(\conv(I,k_1))=\conv(A,k_2)$, then we have
%\[k_1^r=Q^Tk_2^r.\]
%\begin{proof}
%Since $I$ and $k_1$ are of the same size, we have
%\[\conv(I,k_1)=(I^r)^Tk_1^r.\]
%Also notice that $\conv(I,k_1)$ is a number, so $S(\conv(I,k_1))=\conv(I,k_1)$. So we have
%\begin{eqnarray*}
%S(\conv(I,k_1))&=&\conv(A,k_2)\\
%(I^r)^Tk_1^r&=&(QI^r)^Tk_2^r\\
%(I^r)^Tk_1^r&=&(I^r)^TQ^Tk_2^r\\
%\end{eqnarray*}
%By the arbitrary choice of $I$ we have
%\[k_1^r=Q^Tk_2^r.\]
%\end{proof}

For the first case with infinite solutions, we choose the solution with the minimum L2 norm,
\begin{align}
\f_t = \arg\min_{\f_t} \lVert \f_t \rVert_2^2 \hspace{1 em} \text{subject to} \hspace{1 em} \S^T \cdot \f_t = \f_c.
\label{eq:linearp}
\end{align}
The reason that we choose a minimum-norm solution is similar of applying weight decay to the weights, i.e., to reduce over-fitting. A flat filter is likely to have more generalization to various cases. It is easy to get the solution of (\ref{eq:linearp}) by the generalized inverse of $\S^T$,
\begin{align}
\f_t = \S (\S^T \S)^{-1} \cdot \f_c.	\label{eq:fs_case1}
\end{align}

For the second case with no exact solution, we see the problem from a different angle: we take the scaled image $\tilde{\I} = \S \cdot \I$ as an input image, and proximate the original $\I$ with a scaling of $\tilde{\I}$, $\I \approx \tilde{\S} \tilde{\I}$. Here the $\tilde{\S}$ is a scaling-up matrix in the reverted direction of $S$. We turn the Equation~\ref{eq:fc2fs_matrix} into
\begin{align}
\forall \tilde{\I}, \text{~~} \tilde{\I}^T \cdot \f_t = (\tilde{\S} \cdot \tilde{\I})^T \cdot \f_c.
\end{align}
Similar to Equation~\ref{eq:fc2fs_final}, we get
\begin{align}
\f_t = \tilde{\S}^T \cdot \f_c	\label{eq:fs_case2}.
\end{align}

\noindent In our implementation, we use bicubic interpolation ~\cite{keys1981cubic} as the scaling method to transform filters. This method can produce nice scaling results without losing too much information of the original image.

%we denote that by $T^s_{a\rightarrow b}(k)$(or $T^s(k)$ if unambiguous), which means the functions takes in the kernel $k$ of size $a\times a$ and outputs a filter of size $b\times b$.

In our model, we also consider a special scaling operation: horizontal flipping. We add some columns with flipped filters to capture the flipped patterns in input. The scaling matrix for flipping is a symmetric invertible matrix. It is very easy to solve Equation~\ref{eq:fc2fs_final},
\begin{align}
\f_t = \S \cdot \f_c.	\label{eq:fs_case3}
\end{align}

\subsection{Training Multiple Columns}\label{sec:trainmodel}

We integrate all the columns with tied filters into a single model and train them together with back-propagation algorithm. Observing Equation \ref{eq:fs_case1}, \ref{eq:fs_case2} and \ref{eq:fs_case3} in above cases, we find the transformation from canonical filter to any scale is always a linear transformation. That means, the filters in all the columns are tied to the canonical filter by a matrix multiplication,
\[\f_t = \Q\cdot \f_c,\]
where $\Q$ is some transformation matrix. Particularly, $\Q$ is an identity matrix for canonical column. This property makes back-propagation very convenient.
%That is very convenient for back-propagation in training.

%The canonical filter $\f_c$ is multiplied by various transferring matrix to get a filter for each column.
Suppose we have $n$ columns, and the corresponding filters are
\[\f_t^i = \Q_i\cdot \f_c.\]
Define the cost function as $E$, which is a function of all the $\f_t^i$. By the chain rule of derivatives, we get

\begin{align}
\frac{\partial E}{\partial \f_c}
& = \sum_{i=1}^n \left( \frac{\partial \f_t^i}{\partial \f_c} \right)^T \cdot \frac{\partial E}{\partial \f_t^i} \nonumber \\
& = \sum_{i=1}^n \Q_i^T \cdot \frac{\partial E}{\partial \f_t^i}. \nonumber
\end{align}

%If the cost function is $E$, by the chain rule of derivatives we have
%\[\left(\frac{\partial E}{\partial \f_c}\right)^T=\sum_{i=1}^n \left(\frac{\partial E}{\partial \f_t^i}\right)^T\frac{\partial \f_t^i}{\partial \f_c}.\]
%Then we have
%\begin{eqnarray}
%\left(\frac{\partial E}{\partial \f_c}\right)^T&=&\sum_{i=1}^n \left(\frac{\partial E}{\partial \f_t^i}\right)^TQ_i,\nonumber \\
%\frac{\partial E}{\partial \f_c}&=&\sum_{i=1}^n Q_i^T\frac{\partial E}{\partial \f_t^i}.\label{eq:bppro}
%\end{eqnarray}

In training, we first do the back-propagation in each column independently. Then, derivatives of the filters distributed on all columns are transformed and gathered as the canonical filters' derivatives. When the canonical filters are updated by these aggregated derivatives, all the filters on the scaled columns are recomputed by filter transformation from the new canonical filters.

%Thus, the derivative of the cost function with respect to the canonical filter is just the sum of derivatives to all the filters (canonical and transformed) multiplied by the transpose of their transformation matrices. It is thus easy to implement the backpropagation process by which we train the model. After the backpropagation process, we apply the gradient to $\f_c$ and re-compute all $\f_t^i$ for all columns.

\section{Experiment Results}\label{sec:exprRes}

This section presents our experimental results. We begin with a detailed analysis on the scale invariance achieved within the network, followed by end-to-end performance on CIFAR-10 dataset ~\cite{krizhevsky2009learning}. The baseline CNN is close to the Alex network \cite{hinton2012improving}, with 3 layers of convolution. Each layer uses $5\times 5$ receptive size and a stride of 1, pooling of receptive size $3\times3$ and a stride of 2, followed by local normalization. The first convolution is paired with max pooling whereas the latter two is followed by average pooling. {\model} extends it to 6 columns. The first three columns use filter size of 3, 5 and 7, with the last three columns being the flipped versions of the first three. All the weights are regularized and tied to the column with the $5\times5$ non-flipped column. We train these models on standard CIFAR-10 with the training method similar to that in \cite{hinton2012improving}. We use the same hyper-parameters (learning rate, momentum, weight decay) as in \cite{hinton2012improving}. We first train the whole net for 240 epochs, then reduce the learning rate by a factor of ten. We train for another 20 epochs, tune the learning rate again, and train for another 20 epochs to get the final result.

To exploit the invariance property of the model, we need to generate a new test dataset with a mixture of different scales. We crop the central $24 \times 24$ and $28 \times 28$ of the CIFAR-10 images and resize them to $32 \times 32$. This mixed dataset has 3 different scales: small, middle and large. We refer to this dataset as {\it scaled} CIFAR-10 later in this section.

Our experiment results are best viewed in electronic form.

\subsection{Filter Transformation for Scale-Invariance}\label{sec:experiSI}

Consider an arbitrary image $I$ and its scaled version $S(I)$. After applying a filter $f$ and its transformation $T(f)$, their corresponding activations become $\conv(I,f)$ and $\conv(S(I), T(f))$, respectively. As described in Section~\ref{sec:scalefilter}, to achieve scale-invariant pattern matching, we expect the former after scaling is indistinguishable with the latter, i.e.,
\begin{align}
S(\conv(I,f)) \equiv \conv(S(I), T(f)). \label{eq:tarfunc}
\end{align}
In section \secc{scalefilter}, we achieve this for small image patches; in this section, we verify this property for larger images.

Note that the left side of Equation \ref{eq:tarfunc}, $S(\conv(I, f))$ is the scaled activation of the canonical image, and is the design target of our transformation function. So we quantify with relative error using
\[\text{diff}(x, y)=\frac{\lVert x-y \rVert_2}{\lVert x \rVert_2}.\]
where $x = S(\conv(I, f))$ and $y = \conv(S(I), T(f))$.
%\emph{this metrics makes it sensitive to the choice of norm (i.e. if $a$ is too small), how to remove it?}

We compare three different kinds of filter transformation methods. $\trans^I$ is an identity transformation, with which we apply the original filter $f$ onto the scaled image $S(I)$.  $\trans^T$ is the filter transformation described in \secc{scalefilter}. $\trans^S$ is a comparison method, where we directly use simple image sampling  to scale the filter. We also normalize the transformed filter to the same L1 norm as the original filter; we find this normalization performs the best compared to other alternatives. $\trans^S$ takes such a transformation on filter $f$,
\[T(f)=\frac{\lVert f \rVert_1}{\lVert S(f) \rVert_1}S(f).\]

\begin{table}[htbp]
	%\small
	\centering  % 表居中
    \subtable[Scaling up] {
	   \begin{tabular}{lcccc}  % {lccc} 表示各列元素对齐方式，left-l,right-r,center-c
		  \hline
    		Filter &$\trans^I$ & $\trans^S$&$\trans^T$ & pooling \\
            \hline
    		random &11.64\% &1.71\%& 1.67\% &\multirow{3}{*}{11.20\%}\\         % \\ 表示重新开始一行
    		CNN &59.63\% &9.10\% & 9.19\% &\\         % \\ 表示重新开始一行
    		\model &66.11\% &12.88\% &10.94\%& \\         % \\ 表示重新开始一行
		\hline
		\label{tab:scaleupfilter}
    \end{tabular}

    }
    \quad
    \subtable[Scaling down] {
    	\begin{tabular}{lcccc}  % {lccc} 表示各列元素对齐方式，left-l,right-r,center-c
	       	\hline
    		Filter &$\trans^I$ & $\trans^S$&$\trans^T$ & pooling \\
            \hline
    		random &14.14\% &1.81\%& 1.56\% &\multirow{3}{*}{9.22\%}\\         % \\ 表示重新开始一行
    		CNN &85.13\% &20.50\% & 15.85\% &\\         % \\ 表示重新开始一行
    		\model &107.19\% &29.23\% &17.10\%& \\
    		\hline
    	\label{tab:scaledownfilter}
	   \end{tabular}

    }
	\caption{Invariance-by-scaling regarding different filter transformation methods. The smaller the values listed in the table, the better Invariance-by-scaling. }
	\label{tab:exprSI}
\end{table}

We report the filter invariance-by-scaling by measuring
$$\text{diff}(S(\conv(I,f)), \conv(S(I), T(f)))$$
in \tab{exprSI}. We take 100 random images from the test set of CIFAR-10 and take their averaged result. Three canonical filters with size of $5 \times 5$ are considered: random filter (the first row), filters learnt in the baseline CNN model (the second row), and filters learnt in {\model} (the third row). As comparison, non-overlapped $2 \times 2$ max pooling, which is usually considered powerful for scale-invariance, is taken as a non-parametric filter applied to image $I$ and scaled $S(I)$. In Table~\ref{tab:exprSI}(a), we use a scaling-up $S$ that doubles the image size from $32 \times 32$ to $64 \times 64$. Accordingly, $\trans^T$ and $\trans^S$ scale the filter size from $5 \times 5$ to $9 \times 9$. In Table~\ref{tab:exprSI}(b), $S$ is a scaling-down. Image size is halved from $32 \times 32$ to $16 \times 16$, and the filter size is scaled down from $5 \times 5$ to $3 \times 3$.

From Table~\ref{tab:exprSI}, it is clear that convolution with the same filter without any transformation is very sensitive to the image scale (column $\trans^I$). Our filter transformation method (column $\trans^T$) and the sampling-based method (column $\trans^S$) are much more robust to the image scale. $\trans^T$ is almost always better than the simple-minded $\trans^S$, especially when the image is scaled down and we need a more precise filter with a very small size. Considering the fact that $\trans^S$ is hard for back-propagation because of the normalization, our method becomes an apparent choice for transforming filters. Also, when the image is scaled up, our filter transformation is even better than pooling. Considering pooling doesn't need to detect any patterns, it's interesting to see that our method achieves robust invariance-by-scaling. When the image is scaled down, our method is still comparable against pooling. From random filter to {\model} trained one, these filters adapt to some specific scale more and more. Consequently, the invariance-by-scaling of these filters also gets worse (from first row to the last row) as expected.

\begin{figure}[]
	\center
	\includegraphics[angle=0,width=0.5\textwidth]{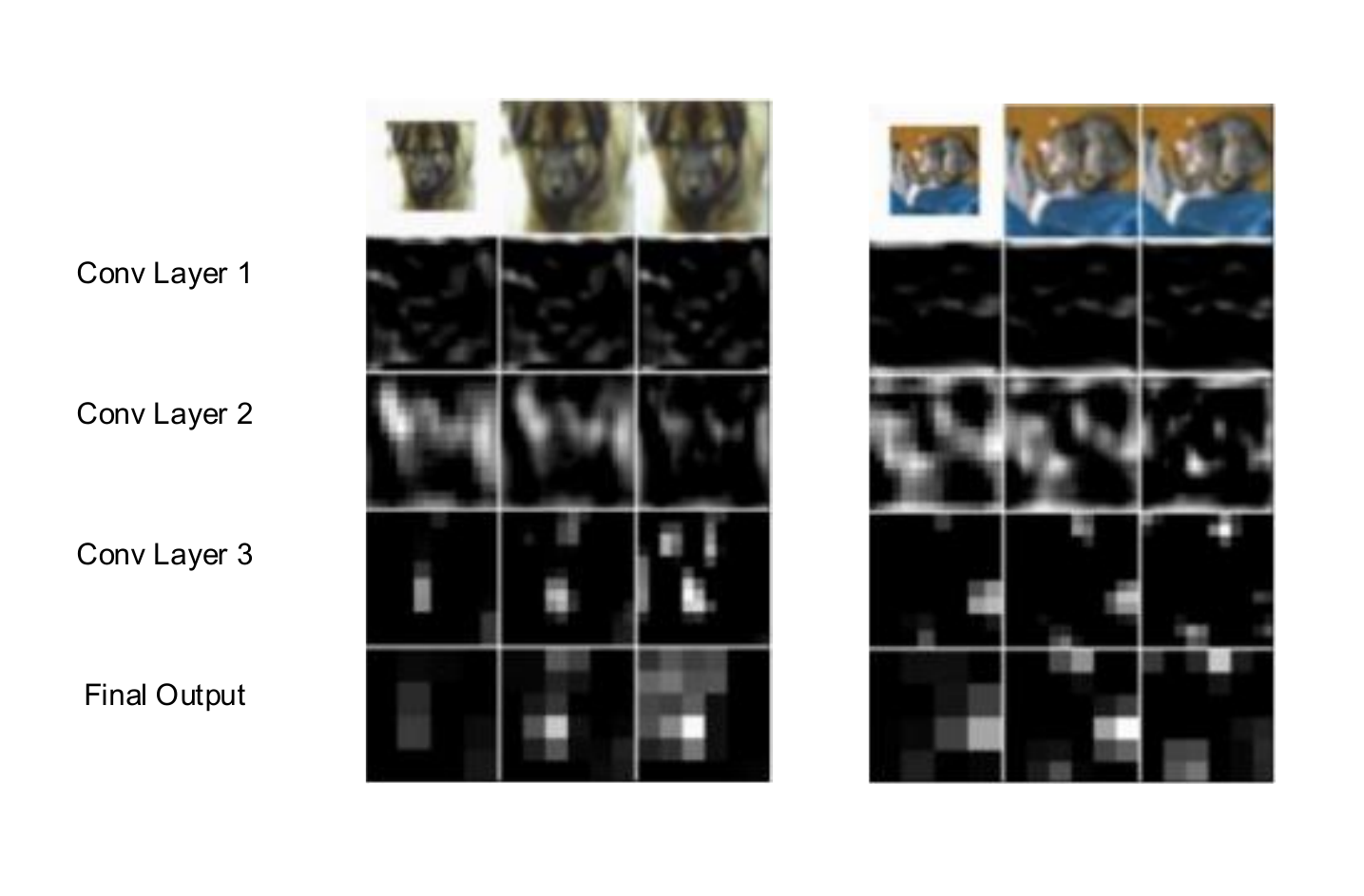}
	\caption{Visualization of activations in each layer. In each of the two examples, the left column shows the original image of size $32 \times 32$ and its activations in each layer using a $5\times5$ filter; the middle and right columns show enlarged image of size $48 \times 48$ and its activations in each layer using the $7\times7$ transformed filter and the same $5\times5$ filter, respectively. The filter is chosen randomly from the {\model} model we trained.}
% generated from the original $5\times 5$ filter. Third column: Same image but activations are generated from a $7\times 7$ filter using our method. Feature maps are resized to a same size for easy comparison.}
	\label{fig:actcompare}
\end{figure}

To give a more concrete feeling of our approach, we inspect the feature maps generated by images of different scales. \fig{actcompare} shows two particular examples, visualizing a feature map in each of the three convolution layers and the final result after pooling and normalization. In each example, the left column is the activations from the original image of $32\times32$, and the other two columns are from scaled image of $48\times48$. The left and middle column are results of using filter of the size $5\times5$ and its transformed $7\times7$ filter. All the feature maps have been scaled to the same size for ease of comparison. From the relatively small difference in each layer, it is clear that applying the transformed filters on the scaled image preserves the essential characteristics of the original. The rightmost column is the result of applying the original $5 \times 5$ filter to the scaled image. It is clear that the fixed filter generates the activations that diverge from that of the original image (the leftmost column) significantly.

\subsection{Multi-Column Features}

\begin{figure*}[]
    \center
    \begin{tabular}{cccc}
        $3 \times 3$ & $5 \times 5$ & $7 \times 7$ & \\
        column 1 & column 2 & column 3 &\\
\includegraphics[width=0.2 \textwidth]{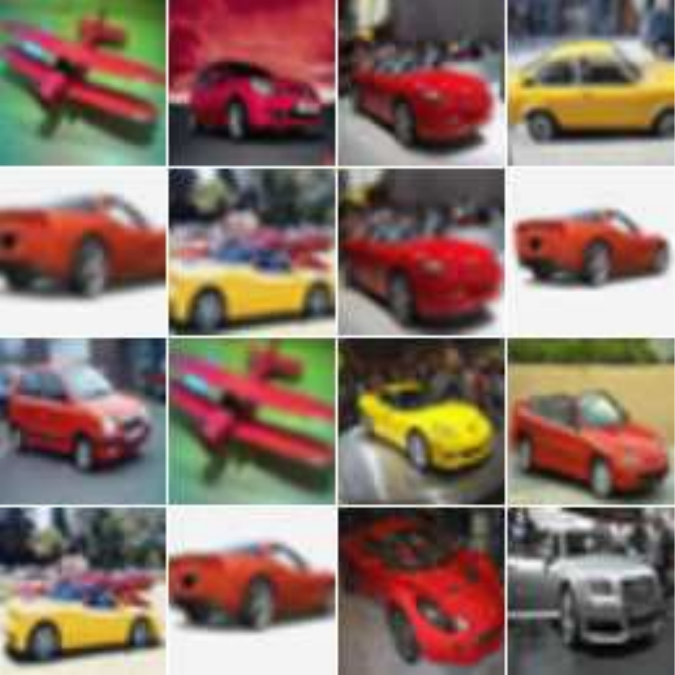} &
\includegraphics[width=0.2 \textwidth]{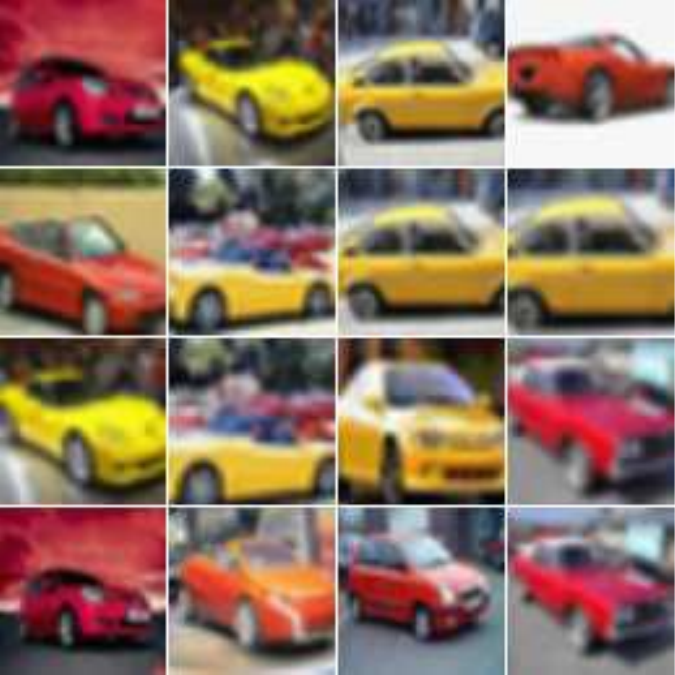} &
\includegraphics[width=0.2 \textwidth]{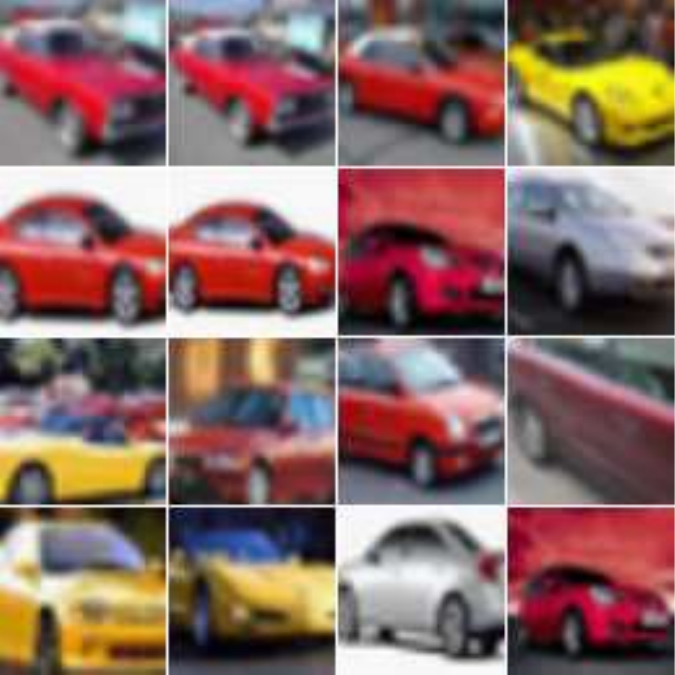} &
\raisebox{7\height}{Non-Flipped} \\
\vspace{0.01 mm}\\
column 4 & column 5 & column 6 & \\
\includegraphics[width=0.2 \textwidth]{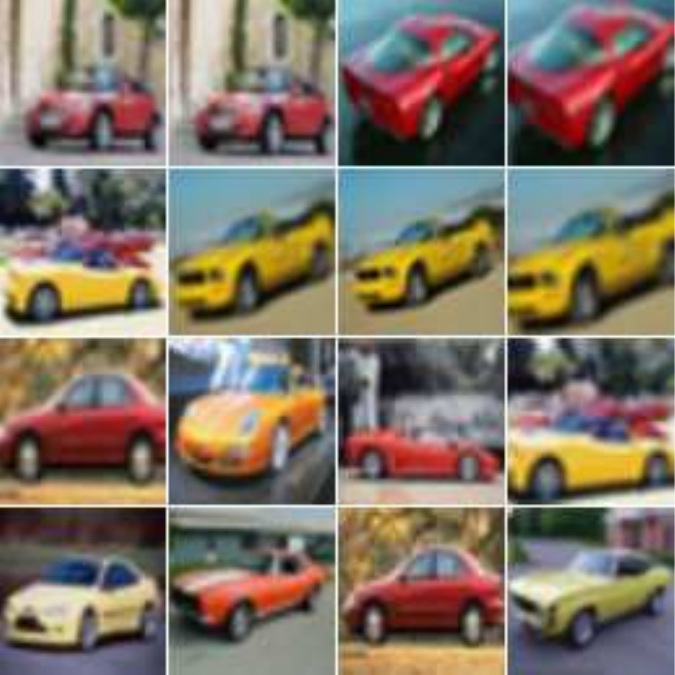}&
\includegraphics[width=0.2 \textwidth]{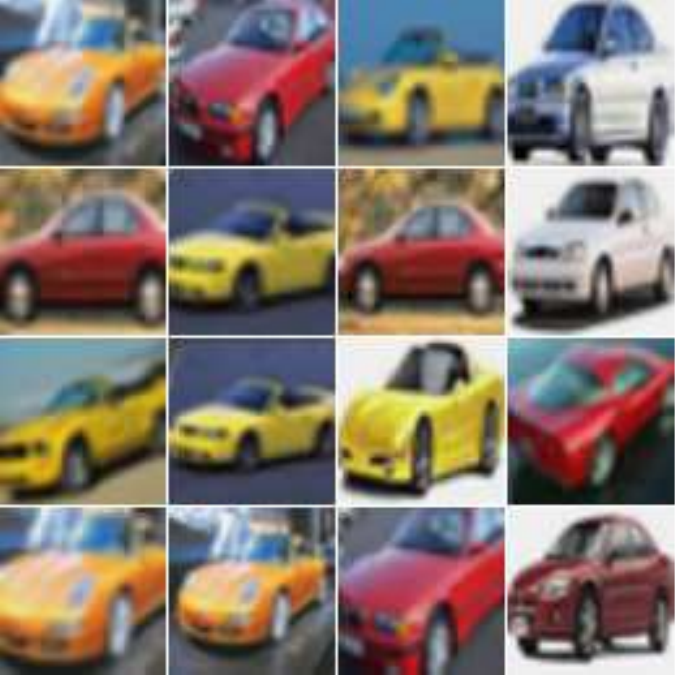} &
\includegraphics[width=0.2 \textwidth]{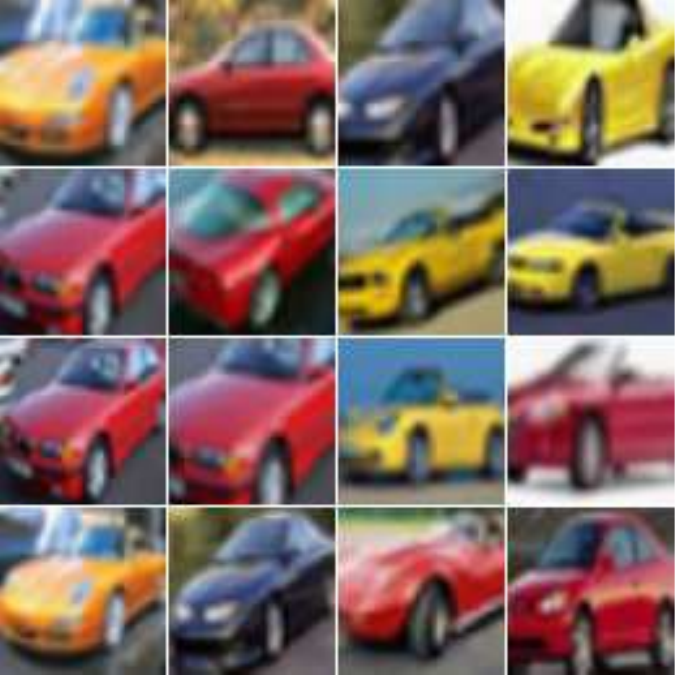} &
\raisebox{7\height}{Non-Flipped} \\
\vspace{0.1 mm}\\
\includegraphics[width=0.2 \textwidth]{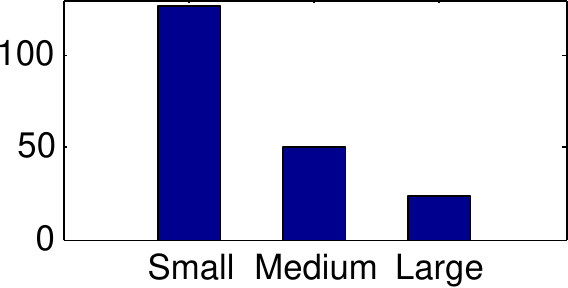}\;\;&
\includegraphics[width=0.2 \textwidth]{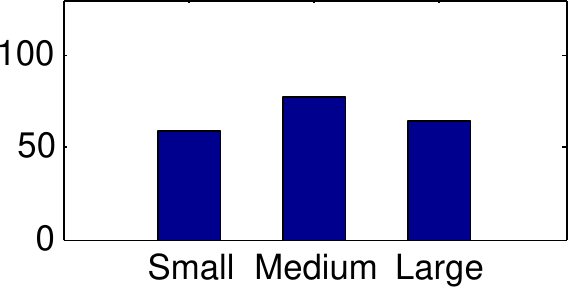}\;\; &
\includegraphics[width=0.2 \textwidth]{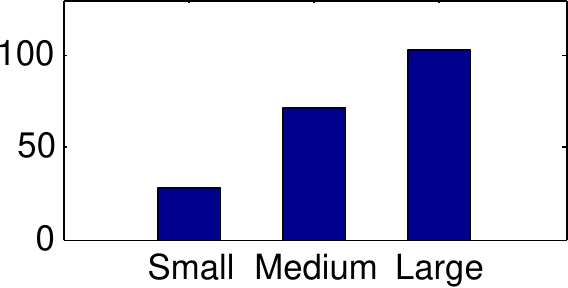}\;\; &
\raisebox{3\height}{Image Scales} \\
    \end{tabular}
\caption{Visualization a filter with respect to the 6 columns. In each column, the top 16 images that causes the largest activation of the filter are shown. The breakdown analysis of top 100 images are shown below each column scale (we combine each two flipped columns together).}
	\label{fig:maxact}
\end{figure*}

To give an idea of what features the different columns learn, we scan over the activations on the last pooling layer caused by 30,000 test images from the scaled CIFAR-10. We randomly pick up a filter in the last layer and visualize the top 16 images that cause the largest outputs of this filter, in each of the 6 columns individually (\fig{maxact}). This method is similar to that used in ~\cite{zeiler2013visualizing}. It can be seen that each column in our model focuses on a particular scale and orientation: the images which causes largest activations get larger from left to right, and the automobiles in the two rows face opposite directions.

In addition to the visual inspection, we try to quantify how sensitive filters of different columns are to the scales. We take the top 100 images that this feature of a given column gets activated the most, then break them down according to which scale they belong to in the data set: small, middle or large. This statics is reported at the bottom of the images in \fig{maxact}. It is clear that columns with small filters ``picks'' the small-scale images more, whereas the columns with larger filters does the opposite.

%To further illustrate \model's ability to detect features at different scales, we plot the activation as a function of the scale of the image. We pick a dog image from the test set of CIFAR-10 and scale it to different sizes(2x larger at most). The scaling is done by cropping the central part of the image. We pick a filter that detects dog features using method similar as in \fig{maxact}, and plot the max activation value at the final layer of each column as a function of the scaling of this image. Results are shown in

When an object for recognition is scaled from small to large, the columns in {\model} will also work in turn to capture this object. We illustrate that in Figure~\ref{fig:corcla}. Using method similar as in \fig{maxact}, we first select a feature map that detects dogs in the last layer. Then we pick a dog image from CIFAR-10 and scale the object into different sizes (2x larger at most).  The max activation value in the feature map are plotted as a function of the object size for each scale column. In Figure~\ref{fig:corcla}, it is clear that when the object is small, the column with $3 \times 3$ filter first captures it and gives a big response. When the object gets larger, activations on this $3 \times 3$-filter column drop. The $5 \times 5$ and $7 \times 7$ columns gradually reach their peak responses in turn. The peaks of the three columns locate with the same interval along the object size, because of the equal-interval filter sizes of 3, 5 and 7. Comparing activation values among different columns is meaningless, because they are eventually summed up with different weights for classification. However, this study clearly shows that by tracing which column is activated more, we can detect a object as well as its scale.

\begin{figure}[]
	\centering
\small
    \begin{tabular}{c}
    \includegraphics[width=0.4 \textwidth]{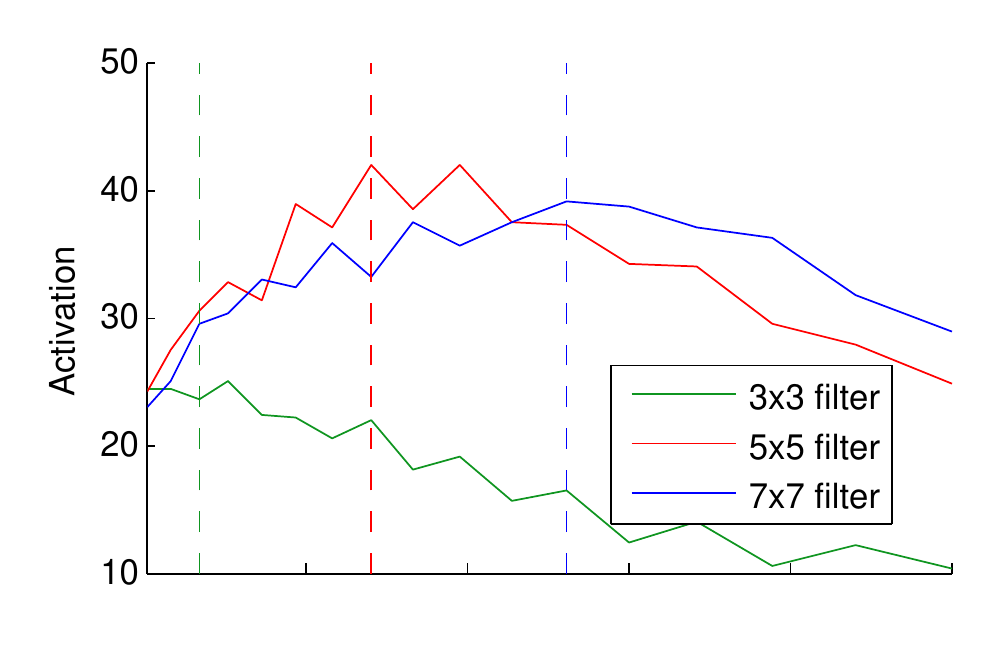}\quad\\
    \includegraphics[width=0.45 \textwidth]{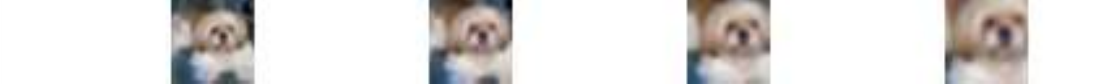}
\end{tabular}
	\caption{Activation versus object size in three columns. Peaks of each curve are illustrated by dash lines.}
	\label{fig:corcla}
\end{figure}

\subsection{Scale-Invariant Classification}\label{sec:scaleinvmod}
%To measure the scale-invariance of the models, we test CNN and SICNN and a changed version of CIFAR, where we keep the original training images but change the testing image to 3 different scales (we do this by taking the central 24x24,28x28,32x2 region of the image and resize them to 32*32).
\begin{table}[htbp]
	\small
	\centering  % 表居中
	\begin{tabular}{c|cc|c}  % {lccc} 表示各列元素对齐方式，left-l,right-r,center-c
		\hline
		Model 			& \tabincell{c}{Standard \\ CIFAR-10}
						& \tabincell{c}{Scaled \\ CIFAR-10}
						& \tabincell{c}{Performance \\ drop}  \\ \hline \hline
		CNN 			& 17.33\% & 24.82\% & 43.22\% \\
		\model			& 14.22\% & 18.83\% & 32.42\% \\ \hline
		Improvement 	& 17.94\% & 24.13\% & \\
		\hline
	\end{tabular}
	\vspace{1 em}
	\caption{Classification error rate, tested on standard CIFAR-10 and scaled CIFAR-10. The last row shows the classification improvement by {\model}. The last column shows the performance drop due to test dataset with more scales.}
	\label{tab:res3scale}
\end{table}

\tab{res3scale} compare the results of the baseline CNN and {\model}, both of which are trained on standard CIFAR-10 dataset. {\model} achieves statistically significant gain on standard CIFAR-10. Its full advantage is more apparent with the scaled CIFAR-10, where CNN has a performance drop of more than $43.22\%$, and {\model} drops by $32.42\%$. We manually examine the error cases, and find that the simple central-crop-resize has cut off many significant features in the scaled CIFAR-10. We speculate {\model} will work better on higher quality multi-scale datasets.
%\emph{I don't understand the use of the last row of the table}
%
%The CNN we use is of a similar structure as the Krizhevsky's CNN stated in ~\cite{hinton2012improving}. We achieved an error rate of 17.44\%, which is slightly worse than the 16.6\% result claimed by Krizhevsky. Seen from the table, easy to see that our model gains more advantage when evaluating on more varied scales, and that the gain from our model is really from the scale-invariance that we incorporate into it.\\
%(TODO:I need to redo the experiment on the net which yields 14.22\% for the column on the right. The current result is of the net of $14.33\%$. Maybe also I need to tune the lr of CNN again.)

%visalization
To verify the above hypothesis, we pick 5 random images in which the object is at the center, and scale them to different sizes; the largest one is the central $16\times 16$ area of the image resized to $32\times 32$. We put these images into both CNN and {\model}, and compare the probability of the correct class. The results are shown in \fig{corcla}. It can be seen that as the scale of the image goes up, the performance of CNN drops whereas {\model} is stable. The only exception among these samples is the one of horse; its scaled up versions start to lose vital features.
\begin{figure*}[]
	\center
    \subfigure[Corresponding Images]{
        \centering
        \includegraphics[width=0.3 \textwidth]{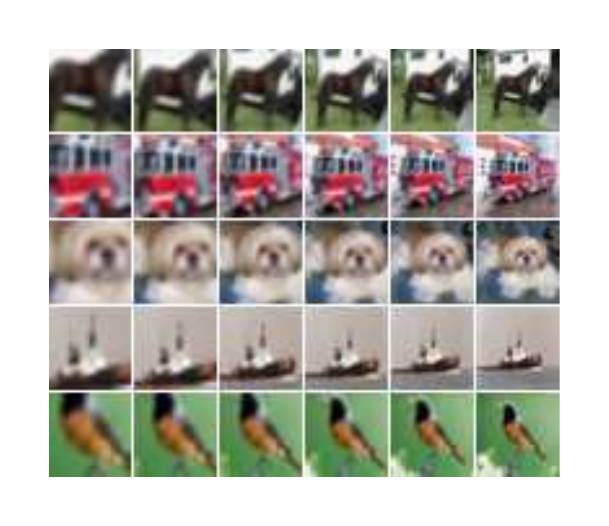}
    }
    \quad
	\subfigure[Result of CNN]{
        \centering
        \includegraphics[width=0.3 \textwidth]{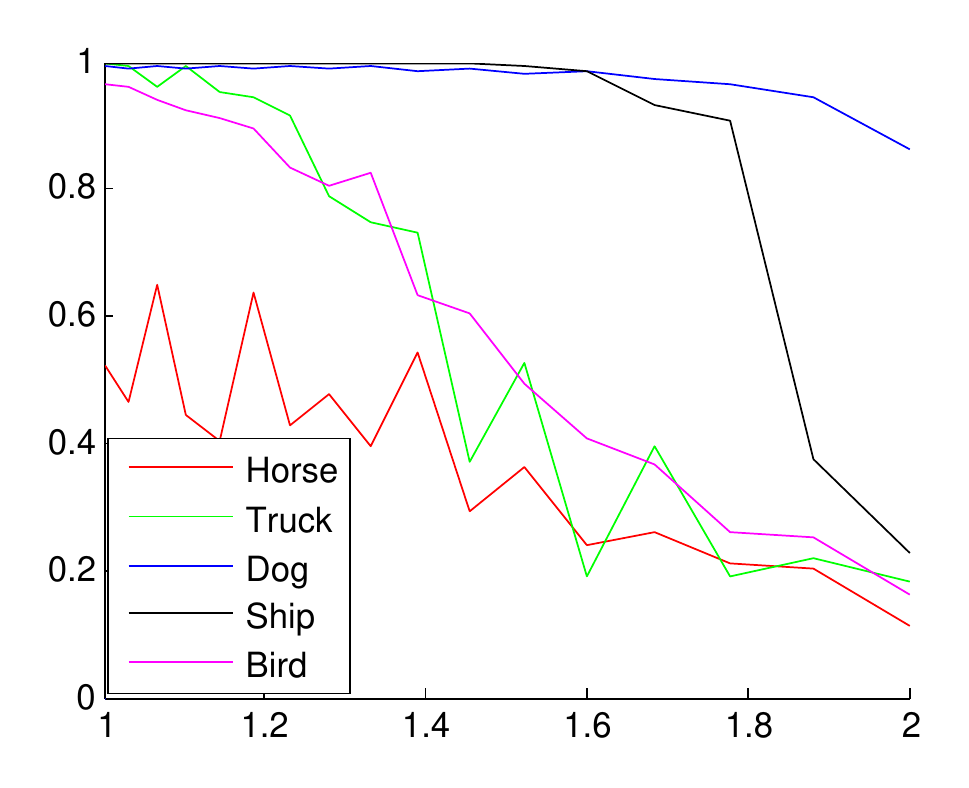}
    }
    \quad
	\subfigure[Result of \model]{
        \centering
        \includegraphics[width=0.3 \textwidth]{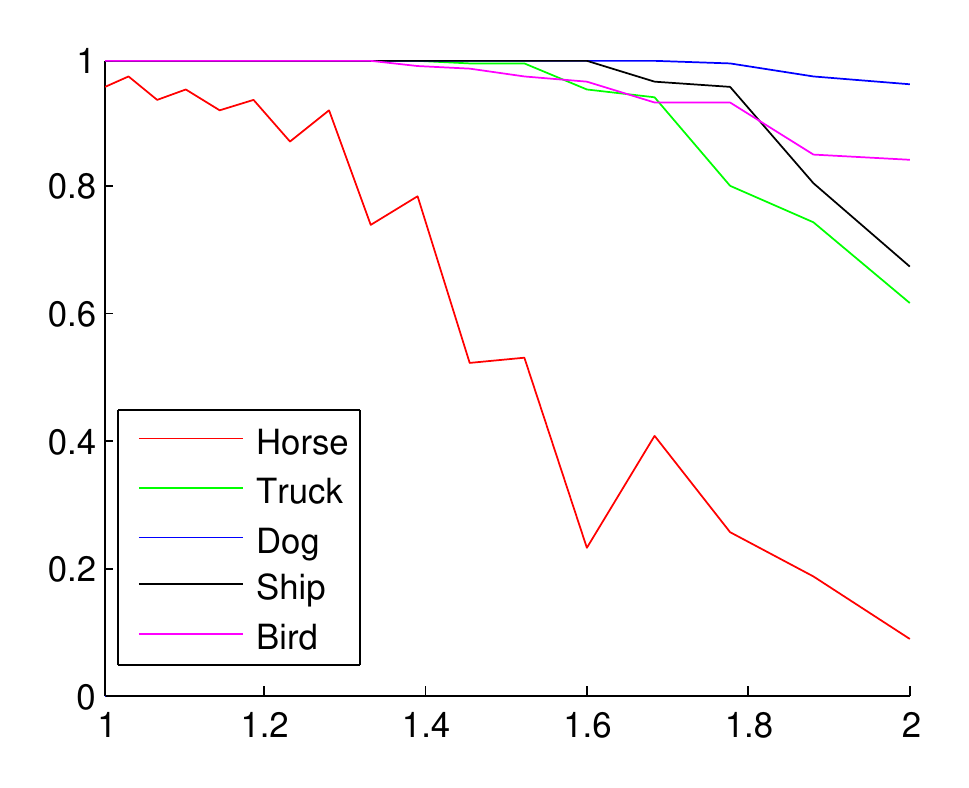}
    }
    \quad
	\caption{Probability of the correct class.}
	\label{fig:corcla}
\end{figure*}

%Like the method applied in ~\cite{gong2014multi,zeiler2013visualizing}, we show the probability that the network output the correct class when we scale the image.

%image retrieval
%\emph{ZZ up to here}. An efficient way of showing what a model has learnt is use image retrieval~\cite{krizhevsky2012imagenet}. For each image, we use the maximum value in all columns for each of the 64 filters as the feature code for each image(i.e., our feature code is only of 64 length). We use the Euclidean distance between vectors to represent similarity between images. The retrieval searches among the 30000 images described in \secc{scaleinvmod}. It can be seen that SICNN is invariant to orientation and scale changes of image that the results include objects facing different directions and of different scales, whereas the CNN produces results of a similar size and similar orientation(\fig{retrieve}).
%\begin{figure*}[!b]
%	\center
%	\includegraphics[width=350px]{resources/retrieval.png}
%	\caption{Retrieve result. First row: 10 random test images from the CIFAR-10 test set. Next 6 row: retrieval result from SICNN. Next 6 row: retrieval result from CNN.}
%	\label{fig:retrieve}
%\end{figure*}

\subsection{Training results on CIFAR-10}

\begin{table}[htbp]
\small
\centering  % 表居中
\begin{tabular}{lc}  % {lccc} 表示各列元素对齐方式，left-l,right-r,center-c
\hline
Method & Testing error\\ \hline  % \hline 在此行下面画一横线
CNN~\cite{hinton2012improving} & 16.6\%\\
CNN+ dropout~\cite{hinton2012improving} & 15.6\%\\
CNN+Spearmint~\cite{snoek2012practical} & 14.98\%\\
{\bf \model} & {\bf 14.22}\%\\
CNN+Maxout~\cite{goodfellow2013maxout} & 11.68\%\\
CNN+Maxout~\cite{goodfellow2013maxout} + {\bf \model} (voting) & {\bf 11.35}\%\\
Maxout-{\bf \model} (2-column) 	& {\bf 11.33}\%\\
Network in Network~\cite{DBLP:journals/corr/LinCY13} & 10.41\%\\
\hline
\end{tabular}
\vspace{1 em}
\caption{Comparison of error rate.}
\label{tab:rescifar}
\end{table}

In \tab{rescifar}, we compare the classification error rate of {\model} with other previous approaches on CIFAR-10~\cite{krizhevsky2009learning}.
We achieve an error rate of 14.22\% on unaugmented data, an improvement of more than 2\% absolute gain over the baseline CNN in \cite{hinton2012improving}. {\model} also exceeds other improvement on CNN, such as dropout \cite{hinton2012improving} and Spearmint~\cite{snoek2012practical}, but is insufficient to catch up with the maxout~\cite{goodfellow2013maxout} and network-in-network \cite{DBLP:journals/corr/LinCY13}, which are the current state of the art. Nevertheless, our method can be combined with these techniques as {\model} is addressing scale-invariance problem, which is a different goal from others. For example, by using an average voting with {\model}, we drop the error rate of maxout model from 11.68\% to 11.35\%. Moreover, by simply adding a extra flipped column to maxout model, we reach an error rate of 11.33\% with a single 2-column maxout-{\model} model. We find these results encouraging, and expect that {\model} will work better on benchmarks that exhibit higher scale variations, as the results in \secc{scaleinvmod} suggest. Replicating SICNN on larger and more complex dataset such as ImageNet is ongoing work.

{\model} takes the form of multi-column CNN without blowing up number of free parameters. As a more direct comparison, we train a 6-column CNN where the filters are independent. Under the same training condition this network suffers severe overfit, the testing error hovers around 19\% while training error already reaches zero.

%which is better than the best result we found achieved by a pure CNN(14.98\%, which uses the Spearmint method~\cite{snoek2012practical}). By using an average voting with the maxout network~\cite{goodfellow2013maxout} which sets the previous state-of-art, we achieve an error rate of 11.35\%. We can't find public code(nor the output results) of NIN~\cite{DBLP:journals/corr/LinCY13}, so we can't obtain the result of voting between NIN and SICNN.
%
%The fact that our method gets improved by voting with maxout indicates that our model learns a different aspect (in our case, the scale invariance) of the CIFAR-10 dataset.\\

\subsection{Incremental Training}\label{incretrain}

\begin{table}[]
\small
\centering  % 表居中
\begin{tabular}{|c|c|c|c|}  % {cccc} 表示各列元素对齐方式，left-l,right-r,center-c
\hline
Model 		& \tabincell{c}{Error on  \\ CIFAR-10}
			& \tabincell{c}{Error on \\ scaled CIFAR-10}
			& cost \\
\hline \hline
CNN & 17.33\% & 24.82\% & 1\\

{\model} & 14.22\% & 18.83\% & 6 \\
\hline
{\model}, {\tt inc-1} & 14.71\% & 20.10\% & 3.5 \\

{\model}, {\tt inc-2} & 16.06\% & 23.24\% & 1 + $\epsilon$ \\
\hline
\end{tabular}
\vspace{1 em}
\caption{Classification error rate and cost for incremental training. The last two rows, {\tt inc-1} and {\tt inc-2}, correspond to the two incremental training methods. All the training costs are normalized to CNN's cost. $\epsilon$ is a very small value. Here $\epsilon = 0.015$.}
\label{tab:resinccnn}
\end{table}

Improving scale invariance does not come for free. In the current configuration, training costs increase linearly with number of columns. Clearly, we can first train one single column,  transform its filters to other columns and finally refine the model. In this ideal setting, it is reasonable to expect that the additional training cost is insignificant.

We explored two incremental training methods. In the first choice (named {\tt inc-1}), we first train a baseline CNN with about half the epochs of a full training, and we build a 6-column {\model} based on the current filters. Then we begin to refine the entire model with the left half of the epochs. In the second choice (named {\tt inc-2}), we continue from a fully trained baseline CNN, and use its filters to build the 6-column {\model}. Then, with all the filter parameters frozen, we only refine the parameters in classifier. As we use a single softmax layer as classifier, the {\tt inc-2} method has a very small extra cost.

Results for incremental learning are summarized in \tab{resinccnn}. Compared with {\model} trained from scratch (the second row), {\tt inc-1} training (the third row) takes nearly half cost, with the model achieving a comparable performance. With the {\tt inc-2} training (the fourth row), although the extra training cost is very small (1.5\%), we still get a model with better performance than baseline CNN. Also by combination with the maxout units ~\cite{goodfellow2013maxout}, we're able to reach an error rate of 11.33\% which is higher than the previous result. Incremental learning does help to balance the performance gain and training efficiency in {\model}.

\section{Conclusion} \label{sec:conclude}

In this paper, we propose a new generalization of CNN, {\model}, where we incorporate scale and flip invariance into the model. This model improves the results of traditional CNN, and complements other optimization techniques. Our results clearly indicate that the model learns the feature in different scales in different columns. The idea is generalizable, can be applied in all aspects where CNN is employed, including supervised and unsupervised learning, recognition, detection, and localization tasks. Our preliminary study also implies a nice trade-off  between performance and training cost. 

Several open problems remain. For example, we can use a different way of summarizing all the columns (instead of concatenation), and different connectivity structure among column (e.g., pair-wise between columns instead of all-to-one against the canonical column). We plan to apply {\model} to larger and more complex datasets such as Imagenet. 
%With this main architecture we can enlarge the model greatly without substantially increasing the parameter number, as suggested by ~\cite{szegedy2014going}, which would be very promising.

{\small
\bibliographystyle{ieee}
\bibliography{bib}
}

\end{document}